%% file: main.tex
\documentclass{article}
\usepackage{times}
\input{math_commands.tex}

\renewcommand{\eqref}[1]{\textup{(\ref{#1})}}
\usepackage[authoryear,round]{natbib}
\setcitestyle{authoryear,round,citesep={;},aysep={,},yysep={;}}
\usepackage{graphicx}
\usepackage{float}
\usepackage{placeins}
\usepackage{caption}
\usepackage{booktabs}
\usepackage{subcaption}
\usepackage{xzc_layout}
\begin{document}
\begin{xzcfrontmatter}
Diffusion Transformers (DiTs) can generate high-quality images and videos, but
generating each sample requires multiple costly DiT forward passes. Two common
ways to accelerate DiT sampling are step distillation, which reduces the number of
sampling steps, and caching, which skips some DiT evaluations by reusing a
tensor computed at an earlier step. Most caching methods decide in advance
which tensor to reuse. After distillation, adjacent sampling steps are farther
apart. Reusing a tensor across this larger gap introduces more error, so
choosing what to cache becomes especially important. We therefore introduce
\textbf{AutoTarget}, a method that chooses the cached tensor for a given model, solver,
and reuse schedule. AutoTarget uses a small set of runs without cache reuse to
measure the error caused by reusing each candidate tensor, then selects the
candidate with the lowest error. We also analyze how an error at one reuse
step affects the final sample. For Euler sampling, we identify cache targets
that produce the same trajectory and show why a stored solver update may not.
Experiments on distilled image and video DiTs show that the best cache target
changes with the model, image resolution, and solver. AutoTarget reduces DiT
evaluations and retained cache storage. Generation quality remains close to
the corresponding uncached run. On the tested PixArt-LCM and FLUX.1-schnell
settings, its calibration ranking matches the ranking from held-out cached
runs. To help others reproduce the method, we provide its core
implementation on GitHub at \url{https://github.com/wali1024-offical/AutoTarget}.
\end{xzcfrontmatter}

\begin{figure}[H]
    \centering
    \includegraphics[width=0.90\linewidth]{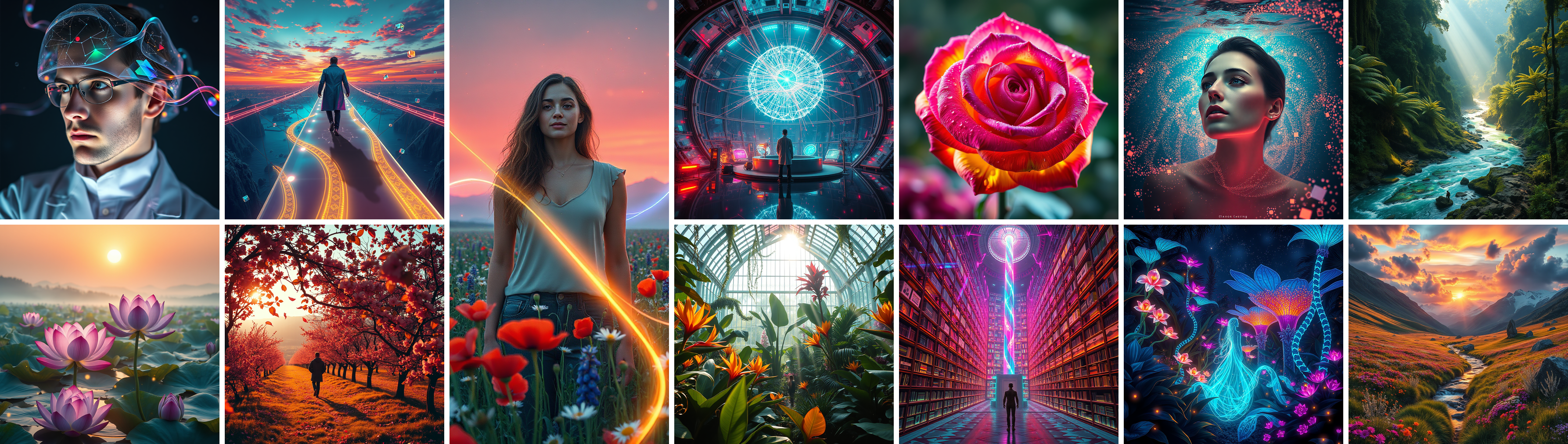}
    \captionsetup{font=footnotesize}
    \caption{Examples generated by AutoTarget with FLUX.1-schnell.}
    \label{fig:teaser}
\end{figure}

\section{Introduction}

Even after step distillation shortens DiT sampling
\citep{salimans2022progressive,song2023consistency,luo2023lcm}, each
remaining forward pass in image and video generation is costly
\citep{chen2024pixartalpha,wan2025,hunyuanvideo2024}. Caching can skip some
of those passes \citep{ma2024deepcache,kahatapitiya2024adacache}, but the
larger gaps between the remaining solver steps make a tensor computed at one
step less reliable when it is reused at a later step.

Many cache methods decide when to reuse a tensor or predict how a stored
feature should change, but keep the tensor type fixed. Different types can
produce different sampling trajectories. For a given model, solver, and reuse
schedule, which tensor should we cache?

\clearpage

\begin{figure}[H]
    \centering
    \includegraphics[width=0.80\textwidth]
    {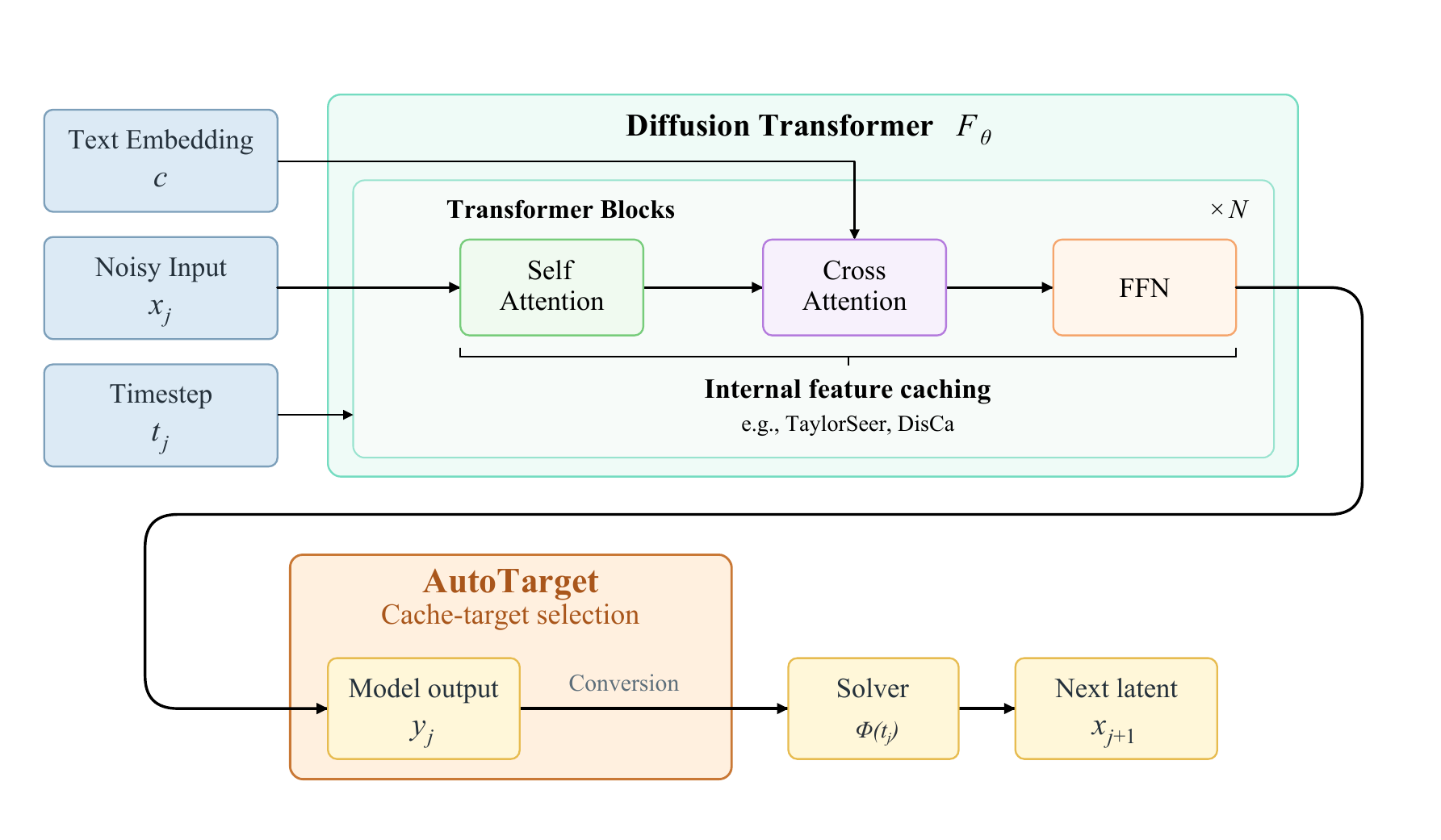}
    \caption{A DiT sampling step. AutoTarget compares the outputs passed to the solver, rather than internal features.}
    \label{fig:dit-cache-flow}
\end{figure}

Figure~\ref{fig:dit-cache-flow} shows the distinction between caching a value
inside the DiT and caching one of its outputs for the solver. We study the
second case. At a reuse step, the stored output is used with a new latent and
solver step size. For a linear-flow model using Euler sampling, a DiT prediction
can be stored as a velocity (the latent's rate of change along the sampling
path), a clean endpoint (the estimated noise-free latent), or a one-step update
(the velocity multiplied by that step's size). These forms describe the same
prediction at the exact step, but need not behave alike when reused later.
Velocity and the clean endpoint give the same trajectory if the endpoint is
converted at each new step. A stored update, however, retains the earlier
step size and can produce a different trajectory when that size changes. This
is why the solver must be part of the cache-target comparison.

With the reuse steps fixed, AutoTarget chooses what to cache using a few
uncached sampling runs. Each run evaluates the DiT at every scheduled step,
allowing us to measure the local reuse error of every valid target. Because
the candidates represent different quantities, we pass each one through its
solver conversion and the frozen VAE decoder before comparing it with the
uncached run. If the two lowest errors are too close, AutoTarget makes no
selection. Otherwise, deployment uses the target with the lowest calibration
error. This selection requires no training and adds no inference-time predictor.
All candidates use
the same number of DiT evaluations and the same solver sequence for a fair
comparison. Figure~\ref{fig:autotarget-workflow} shows the calibration and
deployment stages.

Our contributions are as follows.

\noindent\textbf{Solver-aware cache selection.} AutoTarget chooses what to cache
for a fixed model, solver, reuse schedule, and storage budget using exact
calibration runs.

\noindent\textbf{Reuse analysis.} We show how errors accumulate, when cached forms
give the same Euler trajectory, and why stored updates can differ as step
sizes change.

\noindent\textbf{Results across settings.} PixArt-LCM selects velocity at 512 px
but denoised/clean at 1024 px when only resolution changes
(Figure~\ref{fig:target-switch}). Across image and video DiTs, AutoTarget
saves DiT calls and retained storage while keeping quality near uncached runs.

\section{Related Work}
\label{sec:related}

Modern DiT pipelines build on denoising diffusion and score-based modeling
\citep{ho2020ddpm,song2021score}, latent-space generation
\citep{rombach2022ldm}, and transformer denoisers \citep{peebles2023dit}.
These foundations make the repeated transformer forward pass the main unit
targeted by both step reduction and cache reuse.

\textbf{Few-Step Distillation of Diffusion Models.}
Step distillation compresses the iterative reverse process of diffusion models
into a small number of sampling steps, through progressive distillation
\citep{salimans2022progressive}, consistency training
\citep{song2023consistency,luo2023lcm}, and distribution-matching or
adversarial objectives \citep{yin2024dmd,yin2024dmd2,sauer2024add}. These
recipes yield the distilled image and video DiTs studied in this work,
including PixArt-$\delta$
\citep{chen2024pixartalpha,chen2024pixartdelta}, FLUX.1-schnell
\citep{flusselschnell}, and distilled variants of Wan and HunyuanVideo
\citep{yin2025causvid,huang2025selfforcing,wan2025,hunyuanvideo2024}.
Distillation increases the gap between adjacent solver coordinates. This makes
reuse errors more likely to change the solver trajectory, even though skipping
one DiT evaluation still saves substantial computation
\citep{du2025computeoptimal}. AutoTarget works with a frozen distilled model
and chooses the tensor to cache.

\textbf{Caching and Feature Reuse.}
Caching methods skip full network evaluations by reusing tensors computed at
earlier solver coordinates. Most methods cache predefined internal representations:
deep or block features \citep{ma2024deepcache,wimbauer2024cacheme}, layer
outputs with learned skip routers \citep{ma2024learningtocache},
fast-forwarded blocks \citep{selvaraju2024fora}, polynomially reconstructed
features \citep{feng2025hicache}, attention results
\citep{zhao2025pab,lv2025fastercache}, or non-salient token features
\citep{zou2025toca}; others adapt \emph{when} to reuse via content-adaptive
or timestep-embedding signals
\citep{kahatapitiya2024adacache,liu2025teacache}. Under the larger
inter-step gaps of distilled models, some additionally train predictors that
forecast stale cached features \citep{liu2025taylorseer,zou2026disca}.
Across these methods, the cached representation itself is fixed by design.

\textbf{Cache Target Selection.}
A few recent methods make the cache policy more flexible. They let the model
choose internal features \citep{bu2025dicache}, assign a reuse solver to each
feature dimension \citep{zheng2025hyca}, profile reusable blocks
\citep{ma2025profilingdit}, or optimize the reuse schedule over a full
trajectory \citep{chu2025omnicache,lei2026budcache}. We study a different choice: which tensor
passed to the sampling solver should be cached. The candidates include
velocity, clean and noise endpoints, and solver displacement. AutoTarget groups
candidates with the same solver transitions, then measures local reuse error
for each remaining candidate on exact calibration runs. The chosen target
depends on the model, resolution, and solver; no predictor is trained.

\section{Method}
\label{sec:method}

Given a model, solver, and fixed DiT evaluation schedule, AutoTarget compares
the cache targets supported by the pipeline. It measures their reuse errors
on uncached runs and deploys the lowest-error target on new prompts only if
its error is sufficiently below the runner-up's.

\subsection{DiT Sampling and Cache Targets}
\label{sec:exact-to-cache}

Let $j\in\{0,\ldots,T\}$ be the discrete sampling-step index, and let
$t_0>t_1>\cdots>t_T$ be the corresponding solver coordinates. At each step
$j<T$, the DiT takes the current latent $x_j$, coordinate $t_j$, and
conditioning information $c$ as inputs. Its prediction determines the next
latent through the solver transition:
\begin{equation}
  y_j = F_\theta(x_j,t_j,c), \qquad
  x_{j+1} = \Phi(x_j,y_j,t_j;\eta_j),
  \label{eq:exact-sampling}
\end{equation}
where $F_\theta$ denotes the complete DiT forward pass, $y_j$ is the
prediction consumed by the solver, $\Phi$ is the solver transition, and
$\eta_j$ contains the solver coefficients, history, or other state required
at discrete step $j$. Depending on the pipeline, $y_j$ can be expressed as a
velocity, a clean endpoint, a noise endpoint, or another solver-specific
quantity. We write a clean endpoint as $x^{\mathrm{clean}}$ to distinguish it
from the discrete state $x_0$. These parameterizations arise in diffusion and
flow formulations \citep{karras2022edm,lipman2023flowmatching,liu2023rectifiedflow}.

We call a step that evaluates the full DiT and refreshes the cache an
\emph{anchor}. As shown in Figure~\ref{fig:dit-cache-flow}, a \emph{cache target}
is a tensor obtained at an anchor and converted into the solver's input at a
later step. Reuse skips the DiT evaluation; the conversion rule determines
the stored tensor's effect on the next latent.

For the LCM experiments, we additionally consider a boundary-conditioned
denoised residual. Let $\widehat{x}^{\mathrm{clean}}_j$ be the clean prediction and let
$c_{\mathrm{skip}}(t_j)$ and $c_{\mathrm{out}}(t_j)$ be the original LCM
boundary-condition coefficients. The denoised prediction and residual are
\begin{equation}
  \widetilde{x}^{\mathrm{den}}_j
  =c_{\mathrm{out}}(t_j)\widehat{x}^{\mathrm{clean}}_j
  +c_{\mathrm{skip}}(t_j)x_j,
  \qquad
  d_j=\widetilde{x}^{\mathrm{den}}_j-x_j.
  \label{eq:lcm-denoised-residual}
\end{equation}
At a reuse step, the residual $d_a$ stored at the anchor gives
$\widehat{x}^{\mathrm{den}}_j=x_j+d_a$. The original LCM scheduler then converts
this value to its required input. This residual is different from the Euler
displacement $u_a=h_av_a$ discussed below.
In the PixArt-LCM results, ``denoised/clean'' denotes the numerically checked
group containing $\widetilde{x}^{\mathrm{den}}$ and the predicted clean endpoint
$\widehat{x}^{\mathrm{clean}}$. Its stored representative is the denoised
prediction, which is distinct from the residual $d_a$. The grouping test is
described in Appendix~\ref{app:candidate-construction}.

Let $r$ index a solver-facing cache target considered by AutoTarget. At an
exact anchor step $a$, the candidate is extracted and stored as
\begin{equation}
  m_a^{(r)} = E_r(x_a,y_a,t_a).
  \label{eq:cache-encode}
\end{equation}
Here, $E_r$ extracts or converts the anchor prediction into representation $r$,
and $m_a^{(r)}$ is the stored tensor. At a later reuse step $j$, the corresponding
conversion rule $D_r$ reconstructs the solver input from this tensor:
\begin{equation}
  \widehat{y}_j^{(r)}
  = D_r\bigl(m_a^{(r)},x_j,t_j,\eta_j\bigr).
  \label{eq:cache-decode}
\end{equation}
The rule $D_r$ combines the stored tensor with the current latent, coordinate,
and solver state, for example through an analytic conversion between prediction
parameterizations. This cache conversion is distinct from the VAE decoder used
for calibration below. The original solver then updates the latent:
\begin{equation}
  \widehat{x}_{j+1}^{(r)}
  = \Phi\bigl(x_j,\widehat{y}_j^{(r)},t_j;\eta_j\bigr).
  \label{eq:cached-sampling}
\end{equation}

Let $\mathcal{A}\subseteq\{0,\ldots,T-1\}$, with $0\in\mathcal{A}$, denote the
fixed set of anchor steps. At a step in $\mathcal{A}$, the model
computes a new prediction and refreshes the cache. At every other step, the
full forward pass is skipped and the most recent cached value is reused.

\paragraph{Candidate construction.}
Each pipeline supports a specific set of valid cache targets. We include only
quantities that the frozen pipeline already produces, or can convert exactly,
at an anchor step. These quantities may include the native model output,
velocity, clean and noise endpoints, an LCM denoised residual, and a defined
solver update. Every target has an encoder that specifies what is stored and a
decoder that specifies how the stored value is used later. We omit a target if
its conversion is undefined at any step in the evaluation.

We fix the candidate definitions and grouping criteria before calibration.
Targets with analytically identical solver transitions form one group. For
numerical equivalence, we test the recorded calibration traces using tolerances
fixed in advance (Appendix~\ref{app:candidate-construction}); targets that fail
the test remain separate. We then compare one representative per group. All
candidates use the same data type, tensor shape, anchor steps, and storage size,
as well as the same solver coordinates and coefficients.

\subsection{Equivalent Cache Targets under Euler Sampling}
\label{sec:cache-target-trajectory}

Different representations of a prediction can remain equivalent when cached,
depending on how the solver uses them. For deterministic Euler sampling, let
$h_j=t_j-t_{j+1}>0$ and let $v_j$ be the velocity supplied to the solver. The
latent update is
\begin{equation}
  x_{j+1}=x_j-h_jv_j.
  \label{eq:euler-update}
\end{equation}

For this linear-flow analysis, let $\{t_j\}_{j=0}^{T}\subset[0,1]$ denote the
normalized flow coordinates of the discrete solver steps. The standard linear
interpolation between a clean endpoint and noise is
\begin{equation}
  x_j=(1-t_j)x^{\mathrm{clean}}+t_j\varepsilon,
  \qquad
  v=\varepsilon-x^{\mathrm{clean}},
  \label{eq:linear-flow-path}
\end{equation}
This parameterization gives the following conversions for a velocity prediction
$v_a$ computed at anchor $a$; the inferred endpoints need not equal the true
clean sample and noise:
\begin{equation}
  v_a,
  \qquad
  x_a^{\mathrm{clean}}=x_a-t_av_a,
  \qquad
  \varepsilon_a=x_a+(1-t_a)v_a.
  \label{eq:anchor-targets}
\end{equation}

All three tensors describe the line through $x_a$ in direction $v_a$, although
they store different quantities. We say that two
targets are \emph{equivalent for caching} under a given solver and reuse
schedule if they produce the same latent at every reuse step after a common
anchor.

\paragraph{Proposition 1 (equivalence under Euler sampling).}
Under Eqs.~\eqref{eq:euler-update}--\eqref{eq:anchor-targets}, held velocity,
clean endpoint, and noise endpoint produce the same cached trajectory when
the endpoints are converted at each current state and their decoders are
defined. Between successive anchors, their shared trajectory is
\begin{equation}
  x_j=x_a-(t_a-t_j)v_a.
  \label{eq:held-velocity-trajectory}
\end{equation}
At each reuse step, decoding either endpoint at the current latent recovers
$v_a$. Thus, all three representations give Euler the same velocity and produce
the same next latent. Each cache refresh starts a new segment to which the same
argument applies. Appendix~\ref{app:euler-proof} gives the induction proof.

The proposition assumes that reuse replaces selected DiT evaluations without
changing the solver coordinate sequence. Running the solver only at the anchor
steps uses a different discretization, so the proposition does not cover that
case.

Not every quantity derived at the anchor belongs to this equivalence group.
Suppose we store the Euler update quantity $u_a=h_av_a$ and subtract it directly
from the current latent. Starting from the same $x_j$, this transition differs
from velocity reuse by
\begin{equation}
  \widehat{x}_{j+1}^{(u)}
  -
  \widehat{x}_{j+1}^{(v)}
  =
  (h_j-h_a)v_a.
  \label{eq:fixed-update-mismatch}
\end{equation}
The stored displacement contains the step size $h_a$ from the anchor. If the
current step has a different size, the stored displacement no longer produces
the same update as the three targets above.

\begin{figure*}[!t]
    \centering
    \includegraphics[width=0.96\textwidth]
    {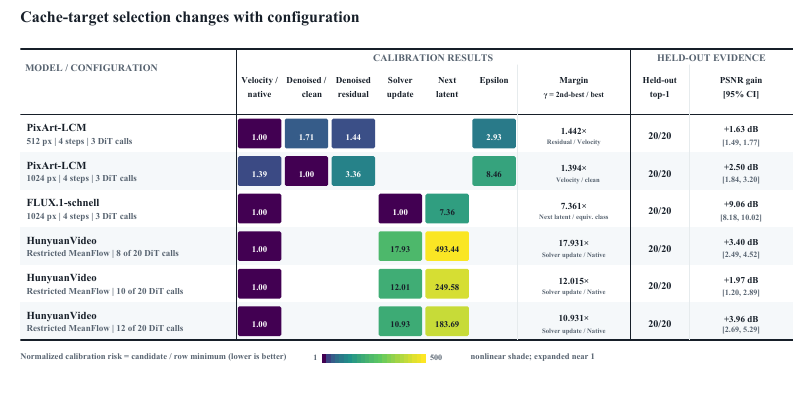}
    \caption{Cache-target selection across configurations. Risks are normalized by the row minimum; held-out columns report top-1 wins and PSNR gains over the runner-up.}
    \label{fig:target-switch}
\end{figure*}

The relationship among targets can change with the solver. Figure~\ref{fig:target-switch}
shows the selected target for each setting and includes a controlled PixArt-LCM
comparison. The two PixArt runs use the same model weights, LCM solver,
four-step coordinate sequence, steps that evaluate the full DiT, prompts, and
random seeds. Only
the image resolution changes. At $512\times512$, AutoTarget selects velocity,
which exceeds the denoised-residual candidate in PSNR by $1.634$ dB on average (95\%
CI $[1.486,1.773]$). At $1024\times1024$, it selects the denoised/clean target,
which exceeds velocity by $2.499$ dB (95\% CI $[1.841,3.196]$). At both
resolutions, the selected target achieves higher PSNR on all 20 held-out prompts.

Cache selection therefore depends on the stored quantity, its conversion at
the current state, and the solver update. AutoTarget compares these choices
under the intended model and sampling configuration.

\subsection{Selecting a Cache Target from Exact Runs}
\label{sec:autotarget-selection}

For every prompt $p$ in a calibration set $\mathcal{P}$, we run the sampler
without cache reuse. At every discrete step, we record the latent
$x_{p,j}^{*}$, model prediction $y_{p,j}^{*}$, and solver state. At a reuse step
$j\notin\mathcal{A}$, let $a=\max\{k\in\mathcal{A}:k<j\}$ be the preceding
anchor in the deployment schedule. Although calibration evaluates every step,
candidate $r$ reuses the value from this scheduled anchor and receives the
exact state at $j$:
\begin{equation}
  \widehat{y}_{p,j}^{(r)}
  =
  D_r\!\left(
    m_{p,a}^{(r)},
    x_{p,j}^{*},
    t_j,
    \eta_{p,j}^{*}
  \right).
  \label{eq:calibration-candidate-decode}
\end{equation}

Every candidate receives the same current state. This design measures the
error introduced at step $j$ without mixing in errors from earlier reuse steps.

Because targets represent different quantities, we compare their effects in a
common output space. Let $Q$ convert the solver input to the latent supplied to
the frozen VAE. At the same exact state, applying this pipeline-specific
conversion to the original and reconstructed inputs gives
\begin{equation}
  z_{p,j}^{*}
  =
  Q\!\left(
    x_{p,j}^{*},
    y_{p,j}^{*},
    t_j;
    \eta_{p,j}^{*}
  \right),
  \qquad
  \widehat{z}_{p,j}^{(r)}
  =
  Q\!\left(
    x_{p,j}^{*},
    \widehat{y}_{p,j}^{(r)},
    t_j;
    \eta_{p,j}^{*}
  \right).
  \label{eq:common-evaluation-latent}
\end{equation}

Using the frozen VAE decoder $G$, we define the calibration risk as
\begin{equation}
  R_{\mathrm{dec}}(r)
  =
  \frac{1}{|\mathcal{P}|\,|\mathcal{J}_{\mathrm{reuse}}|\,D}
  \sum_{p\in\mathcal{P}}
  \sum_{j\in\mathcal{J}_{\mathrm{reuse}}}
  \left\|
    G\!\left(\widehat{z}_{p,j}^{(r)}\right)
    -
    G\!\left(z_{p,j}^{*}\right)
  \right\|_2^2,
  \qquad
  r^{*}
  =
  \arg\min_r R_{\mathrm{dec}}(r).
  \label{eq:decoder-aware-risk}
\end{equation}

Here $\mathcal{J}_{\mathrm{reuse}}=\{0,\ldots,T-1\}\setminus\mathcal{A}$ and $D$
is the number of decoded RGB values per image or video. Thus, $R_{\mathrm{dec}}$
averages squared error over prompts, reuse steps, and all decoded RGB values.
The calculation does not require a separate DiT run
for each candidate or an external quality model. Let $r_1$ and $r_2$ be the
lowest- and second-lowest-risk candidates, and define
$\gamma=R_{\mathrm{dec}}(r_2)/R_{\mathrm{dec}}(r_1)$. The acceptance threshold
was fixed at 1.10 before held-out evaluation. This ratio measures the separation
between the two candidates. AutoTarget deploys $r_1$ only if
$\gamma\geq1.10$; otherwise, it makes no target selection, leaving uncached
sampling or a separately validated policy to the application.

\begin{figure*}[t]
    \centering
    \includegraphics[width=0.96\textwidth]
    {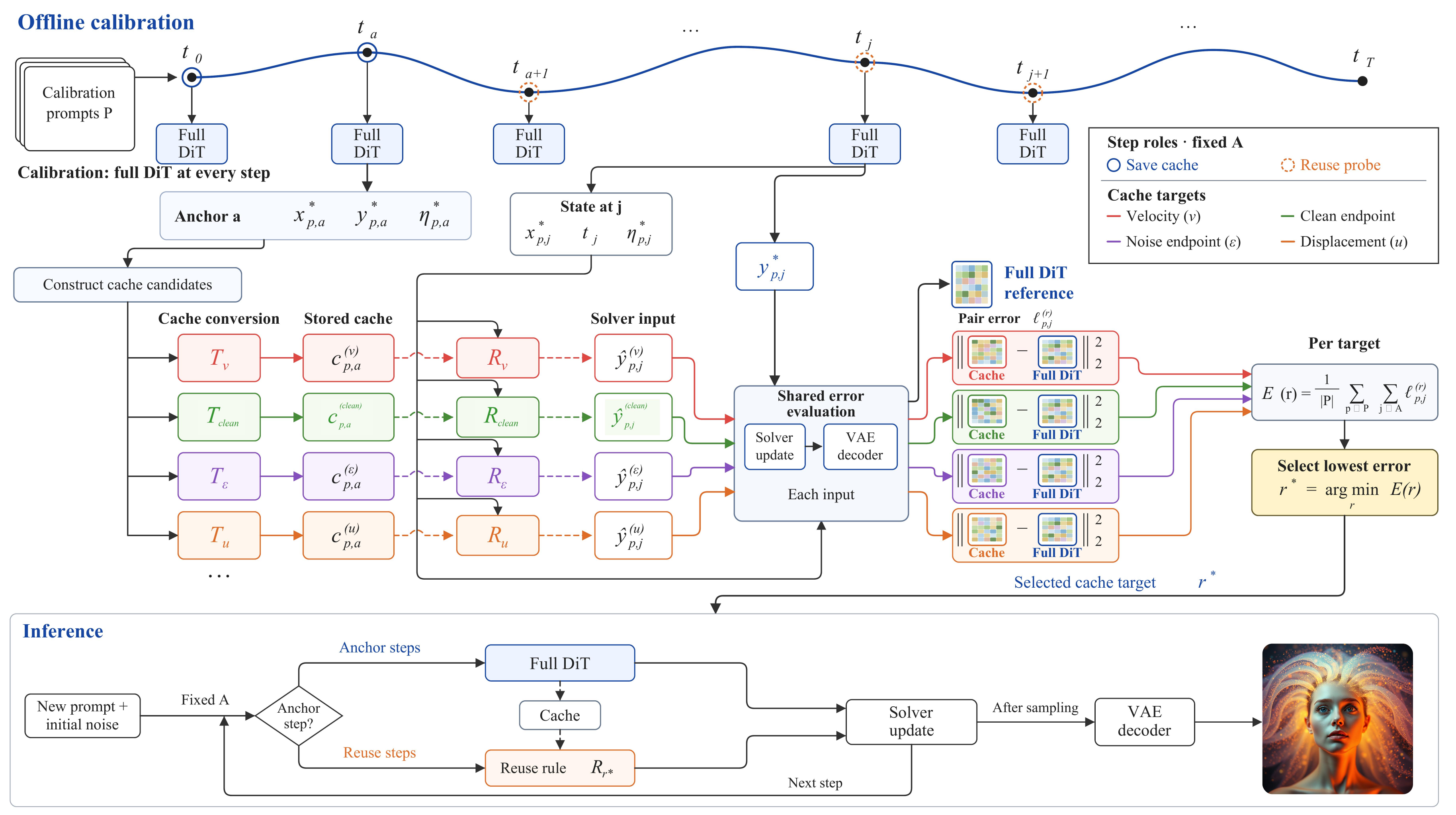}
    \caption{AutoTarget calibrates cache targets on uncached runs and reuses the accepted target on new prompts.}
    \label{fig:autotarget-workflow}
\end{figure*}

Table~\ref{tab:pixart512-calibration-risk} reports the PixArt-LCM calibration
errors. Figures~\ref{fig:calibration-deployment-alignment}--\ref{fig:target-validation}
compare calibration rankings with complete cached runs on held-out prompts,
where reuse errors accumulate. Deployment follows the fixed anchor schedule
with the selected target (Figure~\ref{fig:autotarget-workflow}). Calibration uses
$|\mathcal{P}|T$ DiT evaluations. Converting and VAE-decoding the candidates
happens afterward and requires no additional DiT evaluation.

\paragraph{How local errors accumulate.}
Let $E_j^{(r)}$ measure the accumulated error in the complete state, including
the latent, cache, and solver history. Let $\Delta_j^{(r)}$ measure the error
introduced by one reuse transition from the exact state; it is zero at anchor
steps. If the transitions are locally Lipschitz with factors $\overline{L}_j$
(Appendix~\ref{app:local-global-proof}), then
\begin{equation}
  E_{j+1}^{(r)}
  \leq
  \overline{L}_j E_j^{(r)}
  +
  \Delta_j^{(r)}.
  \label{eq:local-error-recursion}
\end{equation}
Starting from a common initial state, unrolling this recursion gives
\begin{equation}
  E_T^{(r)}
  \leq
  \sum_{j=0}^{T-1}
  \left(
    \prod_{\ell=j+1}^{T-1}
    \overline{L}_{\ell}
  \right)
  \Delta_j^{(r)}.
  \label{eq:global-error-bound}
\end{equation}

The bound relates the final state error to local reuse errors and their
propagation through later steps. AutoTarget ranks candidates using the decoded
RGB error in Eq.~\eqref{eq:decoder-aware-risk}, which provides a common scale
across representations. This empirical score differs from the state error in
the bound; its ranking is tested on held-out cached runs. Appendix~\ref{app:theory}
details this distinction and the derivation.

\section{Experiments}
\label{others}

\subsection{Experimental Setup}
\label{sec:experimental-setup}

We evaluate AutoTarget on few-step image and video DiTs: FLUX.1-schnell,
PixArt-LCM, and a distilled 20-step HunyuanVideo model. Within each few-step
model comparison, methods use the same weights, solver, sampling coordinates,
precision, prompts, and seeds. AutoTarget's candidate targets share a fixed
schedule of full DiT evaluations; baseline methods follow their own reuse
rules, so their DiT call counts can differ.

\paragraph{Evaluation protocol and reporting.}
We evaluate task quality using ImageReward \citep{xu2023imagereward}, CLIP
\citep{radford2021clip}, and VBench \citep{huang2024vbench}, while PSNR, SSIM
\citep{wang2004ssim}, and decoded RGB error measure similarity to a paired
reference. Each cached sample and its reference use the same prompt and initial
noise, controlling these two sources of variation across paired outputs. For
FLUX, the reference is the paired 50-step FLUX.1-dev output
\citep{fluxdev}. For PixArt-LCM, it is the paired
four-step output without caching. Held-out paired comparisons report 95\%
bootstrap confidence intervals.

Overall speedup is measured against the full-step reference, whereas cache-only
speedup isolates the additional acceleration over the matching uncached
few-step sampler. The retained-cache column counts tensors that remain stored
between steps, including any required history. Peak device memory also includes
temporary workspace and is therefore a separate measurement.

Section~\ref{sec:image-results} analyzes the text-to-image results,
Section~\ref{sec:video-results} presents the results on distilled
HunyuanVideo, and Section~\ref{sec:ablation} studies the effects of cache
target selection and calibration size.

\input{tables/image_results.tex}

\input{tables/video_results.tex}

\subsection{Main Results}
\label{sec:main-results}

The full-step FLUX.1-dev and HunyuanVideo rows provide separate references;
the cache baselines are compared within their respective few-step model blocks.

\subsubsection{Text-to-Image Generation}
\label{sec:image-results}

Table~\ref{tab:image-main} shows the effect of reducing FLUX.1-dev from
50 to 17 steps: 17 steps yield $2.67\times$ speedup but reduce ImageReward
from 0.985 to 0.912 and SSIM from 1.0000 to 0.6723. On four-step
FLUX.1-schnell, AutoTarget generates an image in 1.120 seconds with 27.641~dB PSNR and 0.6743
SSIM. Because FLUX.1-schnell is a separate distilled checkpoint, its
$16.00\times$ overall speedup uses 50-step FLUX.1-dev as the reference.
Compared with four-step FLUX.1-schnell without caching, the speedup due to the
cache is $1.289\times$. AutoTarget also obtains ImageReward 0.886 and CLIP
28.376. HiCache obtains a slightly higher ImageReward of 0.896, while AutoTarget
has lower latency and higher PSNR and SSIM, illustrating the different trade-offs
between speed, task quality, and reference fidelity.

PixArt-LCM is more sensitive to the choice of cached tensor. AutoTarget runs in
0.124 seconds, a $1.277\times$ speedup over the same sampler without caching.
Its ImageReward and CLIP scores are 0.413 and 26.070, compared with 0.443 and
26.094 without caching. Its fidelity is 23.15~dB PSNR and 0.7518 SSIM. The
other cache methods obtain 6.16--9.98~dB PSNR and SSIM below 0.18. Although
several baselines preserve task quality on FLUX, their quality and fidelity
drop substantially on PixArt-LCM. This contrast motivates evaluating reuse for
each model configuration; the fixed-schedule ablation below further isolates
the effect of the cached tensor.

AutoTarget also retains less cache data than the compared image baselines:
0.500~MiB on FLUX.1-schnell and 0.125~MiB on PixArt-LCM. The corresponding
baseline ranges are 48--5814~MiB and 18--1512~MiB. These savings follow from
storing a solver-facing tensor instead of the larger feature caches used by
those baselines, and concern retained tensors rather than total inference memory.

\subsubsection{Text-to-Video Generation}
\label{sec:video-results}

On HunyuanVideo (Table~\ref{tab:video-main}), AutoTarget runs in 150.4 seconds,
a $7.55\times$ end-to-end and $1.539\times$ cache-only speedup, with VBench
Semantic/Quality/Total scores of 70.9/80.9/78.9. DisCa obtains
69.9/81.0/78.8 in 151.3 seconds. With only 0.9 seconds separating them, the two
methods have similar latency and overall VBench scores. AutoTarget obtains the
highest Semantic score among the compared cache baselines without requiring a
learned predictor during deployment. The 50-step
reference takes 1136.2 seconds and has a Total score of 78.9. The matching
distilled model without caching takes 231.5 seconds and scores 75.5. The two
speedup columns use these different reference runtimes to distinguish overall
acceleration from the additional gain provided by caching.

\subsection{Ablation Study}
\label{sec:ablation}
Figure~\ref{fig:target-selection-ablation} compares four cache targets on
200 DrawBench prompts with PixArt-LCM at $1024\times1024$. With the model,
solver, and reuse schedule held fixed, denoised/clean achieves the highest PSNR
at 23.155~dB; velocity, denoised residual, and $\epsilon$ achieve 18.340,
12.844, and 10.007~dB, respectively.

To test whether selection depends on the calibration set, we draw subsets of
1, 2, 4, 8, 16, and 32 prompts from 32 recorded calibration runs and recompute
the ranking on 1,000 sampled subsets per size. Every trial selects
denoised/clean, which also has the highest PSNR in the 200-prompt comparison.
Thus, a small calibration set consistently identifies the same target in this
configuration, even as the prompt subset changes.

For further validation, Appendix~\ref{app:reviewer-controls} reports held-out
rankings and calibration errors, together with reduced-step controls,
baseline threshold sweeps, and a comparison of two solvers.

\subsection{Discussion and Limitations}
\label{sec:discussion}

AutoTarget selects a cache target for a fixed schedule of full DiT evaluations.
Changing only PixArt's image resolution reverses the ranking of
velocity and denoised/clean, while the solver comparison changes which
candidates are valid and the order of the lower-ranked targets. These results
show that the preferred target depends on the sampling configuration, even
when the candidate tensors are derived from the same model prediction.

Decoded calibration error provides an empirical ranking whose reliability must
be tested on held-out prompts. The solver-state bound explains error propagation,
but does not establish that this ranking transfers to other prompt distributions.
We therefore fix the selection rule before held-out evaluation and compare
complete cached runs, where errors accumulate across reuse steps. Recalibration
is required when the model, solver, schedule, or precision changes, and each
candidate must have a valid conversion to the solver input. Once calibration
is complete, deployment retains only the accepted tensor and requires no
candidate comparisons, external quality model, or learned predictor.

The shifted LTX-Video setting \citep{hacohen2024ltxvideo} illustrates why the
acceptance rule is needed. Here, $\gamma=1.004$, and the calibration winner
outperforms the alternative on only 55\% of held-out prompts. Always deploying
this target incurs 7.36\% regret relative to the best target for each prompt.
Because the risk ratio falls below the fixed 1.10 threshold, AutoTarget abstains,
allowing the application to retain uncached sampling or a previously validated
cache policy when calibration does not clearly separate the candidates.

\section{Conclusion}
\label{sec:conclusion}

We presented AutoTarget, a method for selecting which tensor to cache in
few-step diffusion transformers. Given a fixed reuse schedule, it compares
candidate targets using uncached sampling runs and deploys the lowest-risk
target when calibration clearly separates it from the runner-up. This procedure
requires no learned predictor and accounts for how the stored tensor is
converted and used by the solver. Our analysis identifies equivalent targets
under Euler sampling and relates local reuse errors to their accumulation
along the sampling trajectory.

Experiments on distilled image and video models show that the preferred target
changes with the sampling configuration. Target selection provides additional
acceleration over uncached few-step sampling with a small cache. Paired
evaluations distinguish reference fidelity from task quality, while held-out
tests assess calibration rankings during complete cached runs. When candidates
are insufficiently separated, AutoTarget declines to select a target, leaving
the fallback policy to the application. These results support treating the
cached tensor as an explicit design choice alongside the reuse schedule,
guided by calibration for the intended model and solver.

\FloatBarrier
\clearpage

\section*{Declaration of AI Assistance}
We used AI-assisted tools to support academic writing, language editing, code
development, debugging, and experiment implementation. The authors reviewed,
verified, and, where necessary, revised all AI-assisted content to ensure its
accuracy and consistency with the research objectives. The authors retain full
responsibility for the scientific content, methodological decisions,
experimental results, and conclusions, as well as the integrity and
reproducibility of this work.

\bibliography{iclr2027_conference}
\bibliographystyle{iclr2027_conference}

\appendix
\section{AutoTarget Procedure}
\label{app:autotarget-procedure}

AutoTarget has two stages: calibration and deployment. Both stages use the same
model, solver, coordinate sequence, and steps for full DiT evaluation. During
calibration, all candidates are tested from the same exact states. Deployment
keeps only the selected target.

\subsection{Candidate Construction}
\label{app:candidate-construction}

At every anchor step, the pipeline provides the model prediction used by the
solver. AutoTarget converts this prediction into the cache targets supported by
the current parameterization. In the controlled Euler example, the candidates
are velocity, the clean endpoint $x^{\mathrm{clean}}$, and the noise endpoint
$\varepsilon$. Another pipeline may also provide a solver update or another
tensor with the same shape as the model output. When candidates are compared
under an equal storage budget, they use the same data type and tensor shape.

For each calibration prompt, an exact run records the latent, model prediction,
coordinate, and solver state at every step. At a reuse step, every candidate is
decoded from the same anchor and receives the same exact current state. The
original solver then converts the candidate into the next latent, or into the
intermediate latent expected by the pipeline. This procedure compares all
candidates at the same point in the computation.

For PixArt-LCM, the boundary-conditioned denoised prediction and predicted
clean endpoint are grouped only if their BF16 one-reuse terminal latents have
relative $L_2$ difference at most $10^{-6}$ and their decoded RGB mean squared
error is at most $10^{-10}$ on every calibration trace. These tolerances were
fixed before calibration. Otherwise, both remain separate candidates. The
denoised prediction represents the group when the checks pass.

\subsection{Selection and Deployment}
\label{app:selection-deployment}

We apply the same pipeline conversion and frozen VAE decoder to the candidate
and exact reference. AutoTarget averages their decoded RGB error over prompts
and reuse steps, then selects the target with the lowest average error. If a
group of targets is analytically identical or passes the numerical check above,
one representative is enough for calibration.

During deployment, the steps assigned to full DiT evaluation compute a new
prediction and refresh the selected tensor. At a reuse step, the selected decoder reads that tensor and
the original solver performs its usual transition. Deployment contains no
branches over candidates, external quality model, or learned predictor. For a
calibration set $\mathcal{P}$ and $T$ sampling steps, the exact runs require
$|\mathcal{P}|T$ DiT evaluations. Once these exact states are available,
testing more candidates requires only conversions and VAE decoding. Those
operations add to the one-time calibration cost but do not require more DiT
evaluations.

In the measured PixArt-LCM calibration, four exact runs take 1.314 seconds,
including text processing. Decoding their exact VAE latents takes 1.528
seconds. Candidate conversion takes 0.059 seconds, and candidate VAE decoding
takes 7.533 seconds. The total is 10.457 seconds, excluding the separately
measured model load of 7.988 seconds. Each later generation saves 0.0688
seconds in this setting, so about 152 generations recover the one-time
calibration cost.

\section{Theoretical Details}
\label{app:theory}

\subsection{Equivalence under Euler Sampling}
\label{app:euler-proof}

Consider one reuse segment beginning at an exact anchor $a$. Let
$h_j=t_j-t_{j+1}>0$ and let $v_j$ denote the velocity consumed by Euler. The
Euler transition is
\begin{equation}
  x_{j+1}=x_j-h_jv_j.
  \label{eq:app-euler-update}
\end{equation}
Under the linear flow convention in Eq.~\eqref{eq:linear-flow-path}, the
anchor prediction can be stored as
\begin{equation}
  x_a^{\mathrm{clean}}=x_a-t_av_a,
  \qquad
  \varepsilon_a=x_a+(1-t_a)v_a.
  \label{eq:app-anchor-endpoints}
\end{equation}
The first reuse state produced by held velocity satisfies
\begin{equation}
  x_j=x_a-(t_a-t_j)v_a.
  \label{eq:app-held-trajectory}
\end{equation}

We prove Eq.~\eqref{eq:app-held-trajectory} by induction over the reuse
steps. At a reuse step $j$, the velocity decoder returns $v_a$. Substituting
Eqs.~\eqref{eq:app-held-trajectory} and~\eqref{eq:app-anchor-endpoints}
into the clean-endpoint decoder gives
\begin{align}
  \frac{x_j-x_a^{\mathrm{clean}}}{t_j}
  &=\frac{x_a-(t_a-t_j)v_a-(x_a-t_av_a)}{t_j} \\
  &=v_a.
  \label{eq:app-clean-decoder}
\end{align}
The noise-endpoint decoder gives the same result:
\begin{align}
  \frac{\varepsilon_a-x_j}{1-t_j}
  &=\frac{x_a+(1-t_a)v_a-[x_a-(t_a-t_j)v_a]}{1-t_j} \\
  &=v_a.
  \label{eq:app-epsilon-decoder}
\end{align}
All three representations provide the same velocity to the Euler transition.
Applying Eq.~\eqref{eq:app-euler-update} gives
\begin{align}
  x_{j+1}
  &=x_j-(t_j-t_{j+1})v_a \\
  &=x_a-(t_a-t_{j+1})v_a,
  \label{eq:app-induction-step}
\end{align}
which establishes the induction step. Every exact refresh begins a new segment
from a common latent, so the same argument applies after each refresh. The
three representations are equivalent for caching under the stated Euler
parameterization and reuse schedule.

For a stored anchor displacement $u_a=h_av_a$, direct reuse gives
$\widehat{x}_{j+1}^{(u)}=x_j-h_av_a$. Reusing the velocity gives
$\widehat{x}_{j+1}^{(v)}=x_j-h_jv_a$. Their difference is
\begin{equation}
  \widehat{x}_{j+1}^{(u)}-\widehat{x}_{j+1}^{(v)}
  =(h_j-h_a)v_a,
  \label{eq:app-update-mismatch}
\end{equation}
which vanishes for that transition when the anchor and current step sizes are
equal.

\subsection{How Local Errors Accumulate}
\label{app:local-global-proof}

Different candidates store different quantities. To analyze them with the same
recursion, we define a complete state for each candidate. This state contains
the latent $x_j$, the cached value $m_j^{(r)}$, and the solver history
$\eta_j^{\mathrm{hist}}$:
\begin{equation}
  \mathcal{Q}_r=\mathcal{X}\times\mathcal{M}_r\times\mathcal{H},
  \qquad
  q_j^{(r)}=(x_j,m_j^{(r)},\eta_j^{\mathrm{hist}}).
  \label{eq:app-lifted-state}
\end{equation}
Let $\Psi^{*,(r)}(\cdot,t_j)$ denote an exact transition on this complete state,
and let $\Psi^{(r)}(\cdot,t_j)$ follow the fixed deployment schedule: it evaluates
the full DiT at anchors and reuses the cache otherwise. The exact reference also
refreshes its auxiliary cache only at anchors, although it evaluates the DiT
at every step. Thus, the two maps agree at anchors, and the reference cache
retains the scheduled anchor value between refreshes. We assume local stability
near the trajectories being compared:
\begin{equation}
  \left\|\Psi^{(r)}(q,t_j)-\Psi^{(r)}(q',t_j)\right\|_r
  \leq \overline{L}_j\left\|q-q'\right\|_r.
  \label{eq:app-lipschitz}
\end{equation}
We measure the error from one cached transition that starts at the exact state:
\begin{equation}
  \Delta_j^{(r)}
  =\left\|
    \Psi^{(r)}(q_j^{*,(r)},t_j)
    -\Psi^{*,(r)}(q_j^{*,(r)},t_j)
  \right\|_r.
  \label{eq:app-local-defect}
\end{equation}
Let $E_j^{(r)}=\|\widehat{q}_j^{(r)}-q_j^{*,(r)}\|_r$ be the accumulated error
at step $j$. Adding and subtracting
$\Psi^{(r)}(q_j^{*,(r)},t_j)$ gives
\begin{align}
  E_{j+1}^{(r)}
  &\leq \overline{L}_jE_j^{(r)}+\Delta_j^{(r)}.
  \label{eq:app-error-recursion}
\end{align}
Starting from a common initial state and unrolling the recursion gives
\begin{equation}
  E_T^{(r)}
  \leq
  \sum_{j=0}^{T-1}
  \left(\prod_{\ell=j+1}^{T-1}\overline{L}_\ell\right)
  \Delta_j^{(r)}.
  \label{eq:app-global-bound}
\end{equation}
This is the error bound used in the main text. It applies to any solver that
satisfies the local stability assumption above.

For a prompt distribution, let
$w_j=\prod_{\ell=j+1}^{T-1}\overline{L}_\ell$ and
$C=(\sum_jw_j^2)^{1/2}$. Cauchy--Schwarz followed by Jensen's inequality gives
\begin{equation}
  \mathbb{E}_p E_T^{(r)}(p)
  \leq
  C\sqrt{\mathbb{E}_p\sum_j\left(\Delta_{p,j}^{(r)}\right)^2}.
  \label{eq:app-expected-bound}
\end{equation}
This bound concerns errors measured in the complete solver state and depends
on the local stability assumption in Eq.~\eqref{eq:app-lipschitz}. AutoTarget
does not estimate its right-hand side: selection uses the decoded RGB error in
Eq.~\eqref{eq:decoder-aware-risk}.

\subsection{Comparing Targets after Decoding}
\label{app:common-evaluation-space}

The norm $\|\cdot\|_r$ measures error in the complete state defined for candidate
$r$. Because different candidates store different quantities, these norms are
not directly comparable. To rank candidates, we use the same conversion $Q$ to
map every prediction to a VAE latent. Eq.~\eqref{eq:decoder-aware-risk} then
compares the decoded RGB outputs from the same frozen VAE. If the VAE decoder
is locally Lipschitz, a change in its input satisfies
\begin{equation}
  \|G(z)-G(z')\|\leq L_G\|z-z'\|.
  \label{eq:app-decoder-lipschitz}
\end{equation}
The decoded error gives one common fidelity measure for all candidates. It is
the score used for selection. The solver-state error in the bound above is a
different quantity.

\section{Experimental Details}
\label{app:experimental-details}

This appendix describes how we obtained the results in
Tables~\ref{tab:image-main} and~\ref{tab:video-main}. Within each model setting,
all methods use the same weights, solver, coordinate sequence, prompts, random
seeds, and numerical precision. Baselines follow their own reuse rules, so
their schedules of full DiT evaluations can differ.

\subsection{Models and Sampling Configurations}
\label{app:models-sampling}

The image experiments use three sampling settings. We run FLUX.1-dev at 50,
25, 20, and 17 steps to measure the effect of reducing the step count. We test
FLUX.1-schnell and the cache baselines with four sampling steps. We also test
PixArt-LCM with four steps and use it for the controlled target comparison in
Figure~\ref{fig:target-selection-ablation}. The video experiments compare a
50-step HunyuanVideo reference with a CFG-distilled 20-step model.

The PixArt-LCM target ablation uses the
\texttt{PixArt-LCM-XL-2-1024-MS} checkpoint with the original LCM scheduler,
$1024\times1024$ resolution, four sampling steps, classifier-free guidance
scale zero \citep{ho2022classifierfree}, and bfloat16 arithmetic. The full DiT runs at steps
$\mathcal{A}=\{0,1,2\}$, so every candidate uses three DiT evaluations and one
reuse step. We measure fidelity on 200 DrawBench prompts
\citep{saharia2022imagen} with paired seeds
beginning at 120000.

The controlled resolution comparison in Figure~\ref{fig:target-switch} uses
the \texttt{PixArt-LCM-XL-2-1024-MS} checkpoint, original LCM scheduler, four
sampling steps, guidance scale 4.5, and bfloat16 arithmetic at both resolutions.
Both resolutions use $\mathcal{A}=\{0,1,2\}$, the same prompt split, and the
same seeds. At $512^2$, calibration uses four prompts with seeds 72000--72003,
and evaluation uses 20 separate prompts. There is one reuse step. The decoded
mean squared error averages over $|\mathcal{P}|=4$ prompts,
$|\mathcal{J}_{\mathrm{reuse}}|=1$ reuse step, and all decoded RGB values.

The table blocks use different fidelity references, as stated in their
captions. The FLUX.1-dev and FLUX.1-schnell blocks compare each output with the
paired 50-step FLUX.1-dev output. For this reason, the FLUX.1-schnell row
without caching does not have infinite PSNR or an SSIM of one. The PixArt-LCM
block uses the paired four-step output without caching as its exact reference.
Within every block, the cached sample and reference use the same prompt and
starting noise.

\subsection{Metrics and Measurement}
\label{app:metrics-measurement}

We measure image quality with ImageReward and CLIP. For video, we report the
Semantic, Quality, and Total components of VBench. The image experiments also
report PSNR and SSIM relative to the paired reference trajectory. Within each
comparison, all latency measurements use the same hardware and execution
settings. \emph{Overall speedup} uses the full-step model as the reference.
\emph{Cache-only speedup} uses the matching few-step model without caching. It
isolates the extra speedup from cache reuse.

The ``Retained cache'' column is computed from the shapes and data types of the
tensors that remain stored between steps, including any required history. Peak
runtime memory is a different measurement because inference can use temporary
workspace that is not retained in the cache.

\subsection{Baseline Configurations}
\label{app:baseline-configurations}

The image baselines are TeaCache, DiCache, TaylorSeer, and HiCache. The video
comparison also includes DisCa. Each baseline uses the cache target and reuse
rule defined by its implementation. On HunyuanVideo, TeaCache uses $l=0.15$,
TaylorSeer uses $N=3,O=1$, and DisCa uses $R=0.2,N=2$. AutoTarget adds no
trained feature predictor. Its controlled target comparisons share an anchor
schedule, and the reduced-step control uses the same number of full DiT
evaluations as the corresponding cached run.

For the PixArt-LCM threshold sweep in
Table~\ref{tab:baseline-sweep}, TeaCache uses thresholds 0.10, 0.30, 0.55,
and 0.80; DiCache uses 0.10, 0.25, 0.391, and 0.60. Every point uses the same
50 prompts, paired seeds, model, solver, and four-step sequence. We report the
average number of full DiT evaluations, abbreviated as FCE. A setting with four
full evaluations performs no cache acceleration.

\raggedbottom
\section{Additional Experimental Results}
\label{app:additional-results}

\subsection{Additional Controls}
\label{app:reviewer-controls}

\begin{figure*}[t]
    \centering
    \includegraphics[width=\textwidth]
    {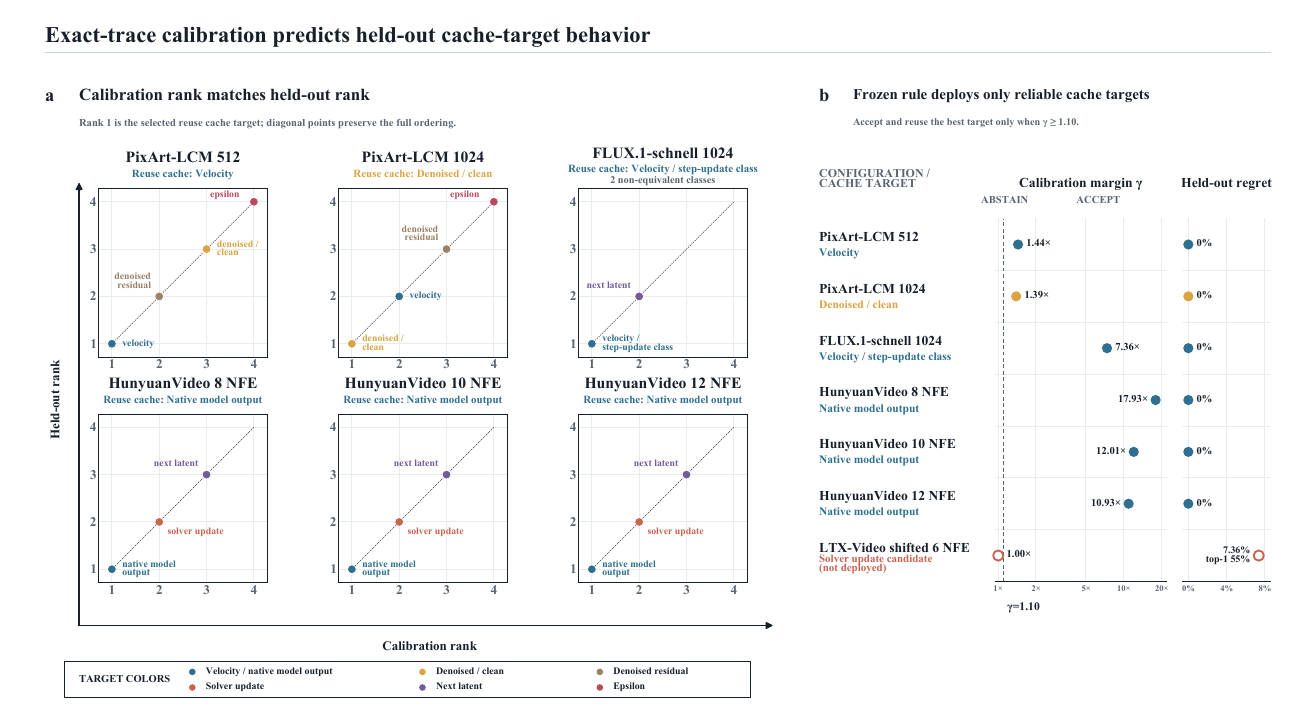}
    \caption{Calibration and held-out cached-run rankings agree in the verified settings. The fixed rule $\gamma\geq1.10$ abstains on shifted LTX-Video.}
    \label{fig:calibration-deployment-alignment}
\end{figure*}

\begin{figure}[H]
    \centering
    \includegraphics[width=0.98\textwidth]
    {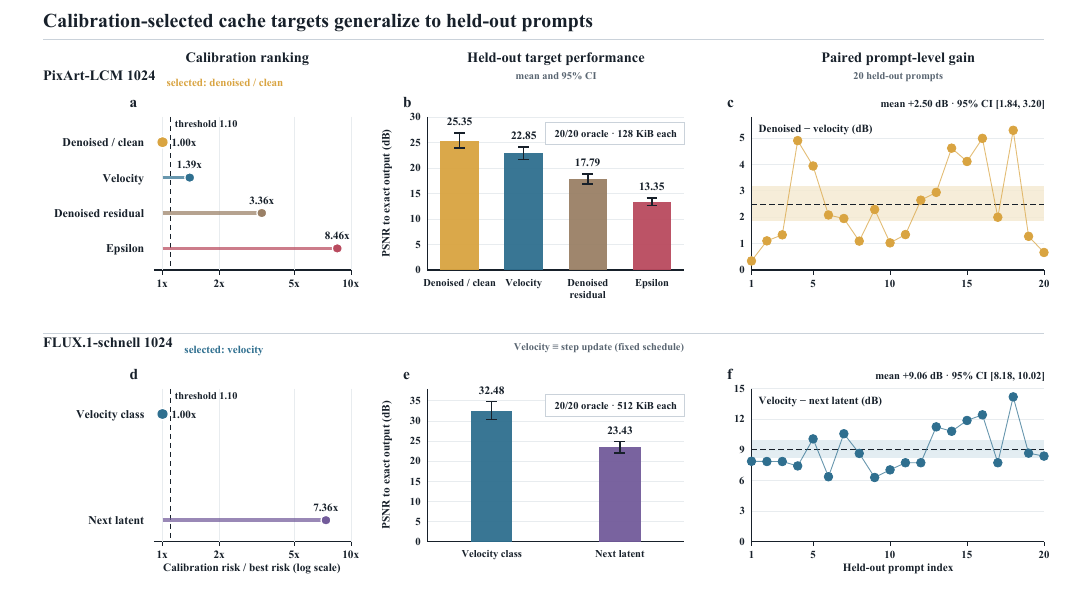}
    \caption{For PixArt-LCM and FLUX.1-schnell, calibration selects denoised/clean and velocity, respectively. Each wins on all 20 held-out prompts; bars and bands show paired 95\% bootstrap intervals.}
    \label{fig:target-validation}
\end{figure}

Table~\ref{tab:pixart512-calibration-risk} reports the four calibration errors
used to select the target at $512^2$.
Figure~\ref{fig:calibration-deployment-alignment} compares the full calibration
ranking with the ranking from held-out cached runs. It also shows the fixed
confidence rule. Figure~\ref{fig:target-validation} gives paired results for
each held-out prompt on PixArt-LCM and FLUX. Table~\ref{tab:reduced-step-control}
compares cache reuse with a solver run that keeps only the same anchor steps. For
PixArt-LCM, it also reports the pipeline's standard three-step schedule.
Table~\ref{tab:baseline-sweep} tests several TeaCache and DiCache thresholds.
Table~\ref{tab:solver-control} compares DDIM \citep{song2021ddim} with
DPM-Solver++ \citep{lu2022dpmsolverpp} on a fixed PixArt-XL model. The solver
changes which targets are valid and reverses the order of the
two lower-ranked targets. In both tested 20-step settings, $\epsilon$ remains
the best target.

\begingroup
\setlength{\intextsep}{8pt plus 1pt minus 1pt}
\input{tables/reviewer_evidence.tex}
\endgroup

\subsection{How the Cache Target Affects Reuse}
\label{app:target-dependent-behavior}

The effect of the cache target differs across models. On FLUX.1-schnell,
several tested cache settings give similar results. On PixArt-LCM, changing the
target produces much larger fidelity differences. On HunyuanVideo, AutoTarget
and DisCa have similar latency and overall VBench scores. These results motivate
selecting a target for each setting. They do not show that AutoTarget is always
best in task quality or speed.

\begin{figure}[H]
    \centering
    \includegraphics[width=0.96\textwidth]
    {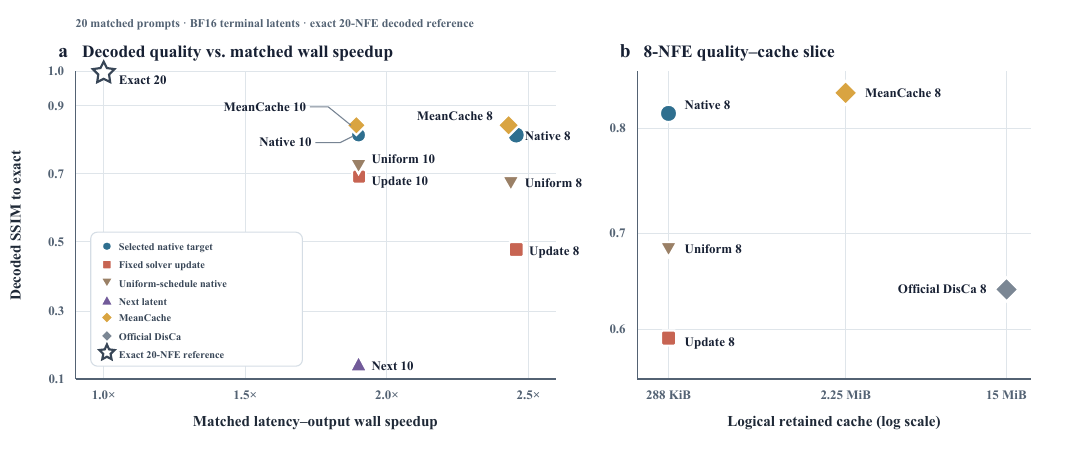}
    \caption{HunyuanVideo speed, fidelity, and retained-cache size on 20 matched prompts. The selected native target retains 288~KiB; cache size excludes temporary runtime memory.}
    \label{fig:hunyuan-pareto}
\end{figure}

\subsection{Task Quality and Trajectory Fidelity}
\label{app:task-fidelity-boundary}

\begin{figure}[H]
    \centering
    \includegraphics[width=0.94\textwidth]
    {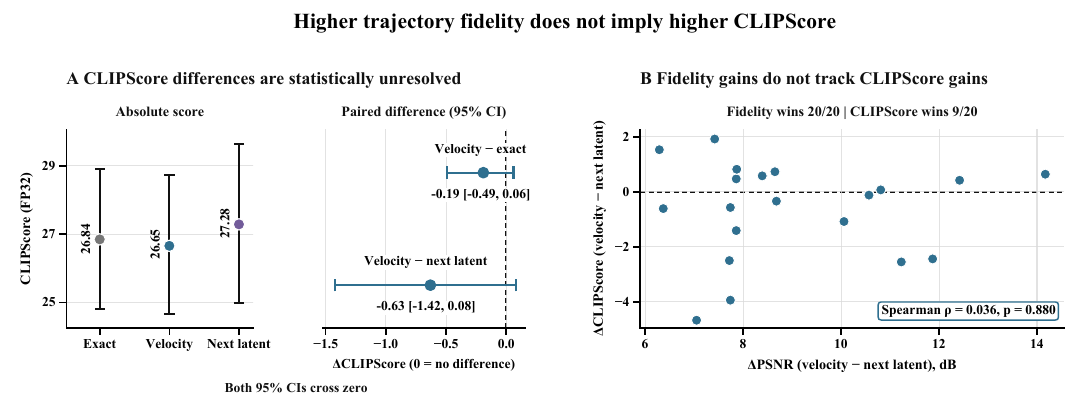}
    \caption{Velocity gives higher fidelity to the exact output than reusing the
    next latent. The paired CLIPScore differences are inconclusive and do not
    follow the PSNR gains.}
    \label{fig:task-fidelity-boundary}
\end{figure}

\newpage
Trajectory fidelity measures similarity to a paired exact output. Prompt--image
alignment measures whether an image matches its text prompt. These are different
properties. In Figure~\ref{fig:task-fidelity-boundary}, velocity has higher
fidelity than next-latent reuse on every held-out prompt. The paired CLIPScore
\citep{hessel2021clipscore} differences are inconclusive and do not follow the
PSNR gains. We report task
quality and reference fidelity separately because neither metric replaces the
other.

\subsection{Calibration Stability}
\label{app:calibration-stability}

\begin{figure}[H]
    \centering
    \includegraphics[width=0.96\textwidth]
    {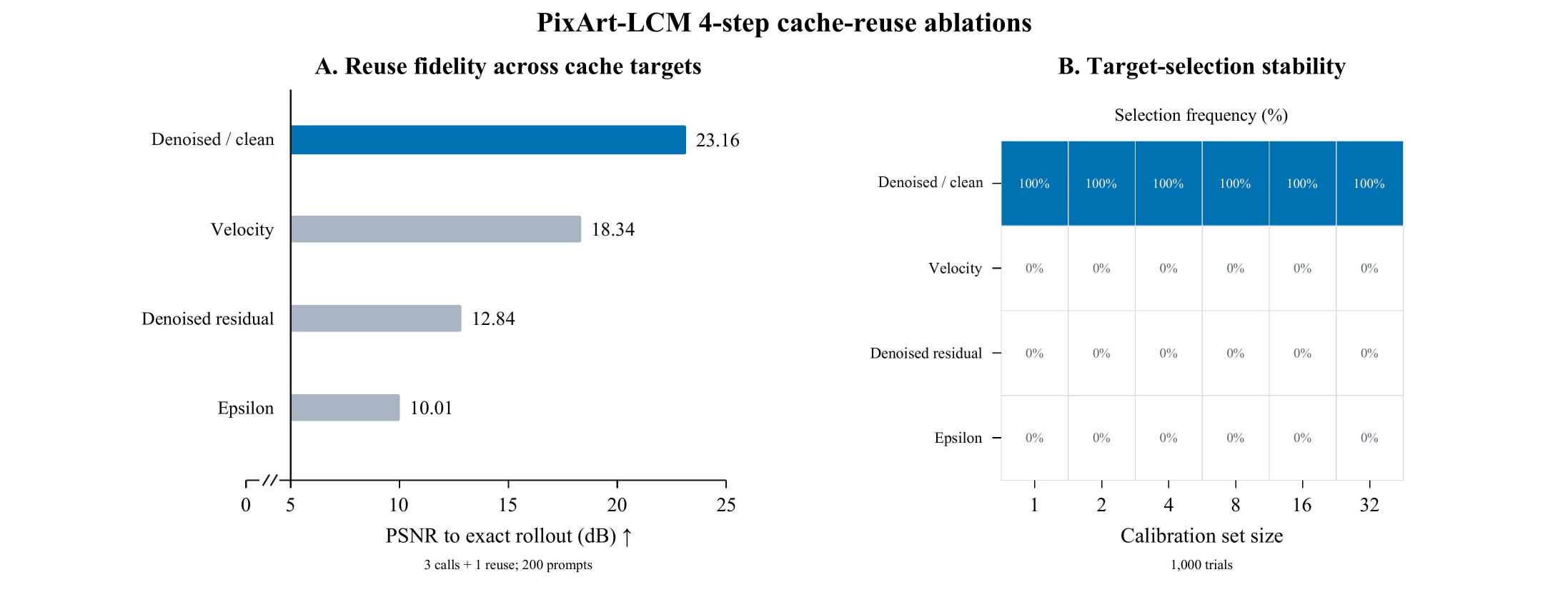}
    \caption{PixArt-LCM target fidelity on 200 DrawBench prompts (left) and selection stability from 32 calibration runs with 1,000 sampled subsets per set size (right).}
    \label{fig:target-selection-ablation}
\end{figure}

Figure~\ref{fig:target-selection-ablation} also varies the number of PixArt-LCM
calibration prompts. Every tested size selects the denoised/clean target, which
also has the highest fidelity in the left panel. In this setting, even a small
calibration set is enough to identify the same target. The analysis starts from
32 recorded calibration runs. For each set size in
$\{1,2,4,8,16,32\}$, we draw 1,000 subsets without replacement using the fixed
random seed 9142026 and recompute the ranking. The 200 prompts in the target
comparison belong to the left panel. The 1,000 subset trials belong to the
calibration-size experiment in the right panel.

These additional results use the same protocol as the main tables. They explain
the reported results in more detail and do not add a new benchmark, metric, or
comparison setting.

\section{Scope and Limitations}
\label{app:scope-boundary}

The general analysis applies to a discrete sampling solver when its transition
on the complete state satisfies the local stability condition in
Eq.~\eqref{eq:app-lipschitz}. For Euler sampling with the linear-flow
parameterization, we give a closed-form proof of which targets are equivalent.
For other solvers, AutoTarget can construct and compare candidates in the same
way, but their ranking must be measured from that solver's transitions.

The calibration error applies to the tested model, solver, coordinate sequence,
and reuse schedule. It does not guarantee the same ranking after any of these
choices changes or on every prompt distribution. The decoded VAE error is the
score used to compare targets in practice. The theoretical bound measures error
in a state that also contains the cache and solver history. Our empirical and
theoretical claims concern these two different quantities.

Cache reuse and step removal are different operations. In our control, both
methods run the DiT at the same anchor steps. They use the same number of DiT
evaluations. Cache reuse still
executes every transition in the original four-step solver sequence. The
reduced-step run executes the solver only at the anchors. Their fidelity is
nearly equal for the tested anchors. By comparison, the standard three-step
PixArt-LCM schedule is much less faithful to the four-step reference. These
results apply only to the tested schedules and solvers.

\end{document}

%% file: math_commands.tex
\usepackage{amsmath,amsfonts,bm}

\def\eqref#1{equation~\ref{#1}}

\def\1{\bm{1}}

\DeclareMathAlphabet{\mathsfit}{\encodingdefault}{\sfdefault}{m}{sl}
\SetMathAlphabet{\mathsfit}{bold}{\encodingdefault}{\sfdefault}{bx}{n}



%% file: tables/image_results.tex

\begin{table*}[t]
\centering
\caption{Text-to-image results. Overall speedup is relative to 50-step FLUX.1-dev; cache-only speedup is relative to each uncached few-step model. Fidelity uses paired 50-step FLUX.1-dev outputs for FLUX and paired uncached four-step outputs for PixArt-LCM.}
\label{tab:image-main}
\scriptsize
\setlength{\tabcolsep}{3pt}
\renewcommand{\arraystretch}{1.28}

\resizebox{\textwidth}{!}{%
\begin{tabular}{lcccccccc}
\toprule
Method
& \multicolumn{3}{c}{Efficiency}
& \multicolumn{2}{c}{Image Quality}
& \multicolumn{2}{c}{Paired-output Fidelity}
& Retained cache \\
\cmidrule(lr){2-4}
\cmidrule(lr){5-6}
\cmidrule(lr){7-8}

& Latency (s)$\downarrow$
& Overall$\uparrow$
& Cache-only$\uparrow$
& ImageReward$\uparrow$
& CLIP$\uparrow$
& PSNR (dB)$\uparrow$
& SSIM$\uparrow$
& MiB$\downarrow$ \\
\midrule

\multicolumn{9}{c}{\textbf{FLUX.1-dev step-budget reference}} \\
\midrule

FLUX.1-dev (50 steps)
& 17.92
& 1.00$\times$
& \textit{N/A}
& 0.985
& 32.187
& $\infty$
& 1.0000
& 0.00 \\

FLUX.1-dev (25 steps)
& 9.31
& 1.93$\times$
& \textit{N/A}
& 0.967
& 32.041
& 29.431
& 0.7315
& 0.00 \\

FLUX.1-dev (20 steps)
& 8.14
& 2.20$\times$
& \textit{N/A}
& 0.946
& 30.912
& 29.037
& 0.6991
& 0.00 \\

FLUX.1-dev (17 steps)
& 6.71
& 2.67$\times$
& \textit{N/A}
& 0.912
& 29.308
& 28.795
& 0.6723
& 0.00 \\

\midrule

\multicolumn{9}{c}{\textbf{FLUX.1-schnell}} \\
\midrule

FLUX.1-schnell (4 steps)
& 1.444
& 12.41$\times$
& 1.000$\times$
& 0.851
& 28.159
& 27.632
& 0.6625
& 0.00 \\

TeaCache
& 1.446
& 12.39$\times$
& 0.999$\times$
& 0.851
& 28.159
& 27.632
& 0.6321
& 48.00 \\

DiCache
& 1.131
& 15.85$\times$
& 1.277$\times$
& 0.852
& 28.205
& 25.484
& 0.6422
& 192.96 \\

TaylorSeer
& 1.151
& 15.57$\times$
& 1.255$\times$
& 0.879
& 28.222
& 26.347
& 0.6412
& 3876.00 \\

HiCache
& 1.229
& 14.58$\times$
& 1.175$\times$
& \textbf{0.896}
& 28.349
& 26.986
& 0.6577
& 5814.00 \\

\textbf{AutoTarget}
& \textbf{1.120}
& \textbf{16.00$\times$}
& \textbf{1.289$\times$}
& 0.886
& \textbf{28.376}
& \textbf{27.641}
& \textbf{0.6743}
& \textbf{0.500} \\

\midrule

\multicolumn{9}{c}{\textbf{PixArt-LCM}} \\
\midrule

PixArt-LCM (4 steps)
& 0.158
& \textit{N/A}
& 1.000$\times$
& 0.443
& 26.094
& $\infty$
& 1.0000
& 0.00 \\

TeaCache
& 0.127
& \textit{N/A}
& 1.240$\times$
& -2.257
& 17.218
& 7.04
& 0.0517
& 18.00 \\

DiCache
& 0.134
& \textit{N/A}
& 1.175$\times$
& -1.916
& 16.983
& 6.16
& 0.1775
& 54.00 \\

TaylorSeer
& 0.145
& \textit{N/A}
& 1.088$\times$
& -1.933
& 23.045
& 9.41
& 0.1047
& 1512.00 \\

HiCache
& 0.148
& \textit{N/A}
& 1.058$\times$
& -1.793
& 23.185
& 9.98
& 0.1161
& 1512.00 \\

\textbf{AutoTarget}
& \textbf{0.124}
& \textit{N/A}
& \textbf{1.277$\times$}
& \textbf{0.413}
& \textbf{26.070}
& \textbf{23.15}
& \textbf{0.7518}
& \textbf{0.125} \\

\bottomrule
\end{tabular}%
}
\end{table*}

%% file: tables/video_results.tex

\begin{table*}[t]
\centering
\caption{HunyuanVideo results. Overall and cache-only speedups are relative to the 50-step and uncached CFG-distilled 20-step models, respectively.}
\label{tab:video-main}
\scriptsize
\setlength{\tabcolsep}{5pt}
\renewcommand{\arraystretch}{1.2}

\resizebox{\textwidth}{!}{%
\begin{tabular}{lcccccc}
\toprule
Method
& Latency (s)$\downarrow$
& Overall$\uparrow$
& Cache-only$\uparrow$
& VBench Semantic$\uparrow$
& VBench Quality$\uparrow$
& VBench Total$\uparrow$ \\
\midrule

HunyuanVideo (50 steps)
& 1136.2
& 1.00$\times$
& \textit{N/A}
& 72.4
& 80.5
& 78.9 \\

HunyuanVideo (CFG-distilled, 20 steps)
& 231.5
& 4.91$\times$
& 1.000$\times$
& 63.2
& 78.6
& 75.5 \\

TeaCache ($l=0.15$)
& 234.8
& 4.84$\times$
& 0.986$\times$
& 64.5
& 79.3
& 76.3 \\

DiCache
& 302.8
& 3.75$\times$
& 0.765$\times$
& 58.8
& 75.6
& 72.2 \\

TaylorSeer ($N=3$, $O=1$)
& 265.0
& 4.29$\times$
& 0.874$\times$
& 64.3
& 79.4
& 76.4 \\

HiCache
& 260.5
& 4.36$\times$
& 0.889$\times$
& 64.9
& 79.6
& 76.7 \\

\textbf{DisCa} ($R=0.2$, $N=2$)
& 151.3
& 7.51$\times$
& 1.530$\times$
& 69.9
& \textbf{81.0}
& 78.8 \\

\textbf{AutoTarget}
& \textbf{150.4}
& \textbf{7.55$\times$}
& \textbf{1.539$\times$}
& \textbf{70.9}
& 80.9
& \textbf{78.9} \\

\bottomrule
\end{tabular}%
}
\end{table*}

%% file: tables/reviewer_evidence.tex

\begin{table}[H]
\centering
\caption{PixArt-LCM $512^2$ calibration errors for the controlled setting in Figure~\ref{fig:target-switch}. Each value is the mean squared decoded RGB error over four prompts and one reuse step. Velocity has the lowest error. The ratio of the second-lowest error to the lowest is $\gamma=1.442$.}
\label{tab:pixart512-calibration-risk}
\scriptsize
\setlength{\tabcolsep}{6pt}
\renewcommand{\arraystretch}{1.14}
\begin{tabular}{clcc}
\toprule
Rank & Candidate target & Calibration risk & Risk / selected \\
\midrule
1 & velocity & \textbf{0.005963} & 1.000 \\
2 & denoised residual & 0.008601 & 1.442 \\
3 & denoised/clean & 0.010180 & 1.707 \\
4 & $\epsilon$ & 0.017475 & 2.931 \\
\bottomrule
\end{tabular}
\end{table}

\begin{table}[H]
\centering
\caption{Reduced-step controls with the same number of DiT evaluations.
``Anchor-only'' runs the solver only at the steps that refresh the cache.
``Default three-step'' uses the standard three-step schedule from the LCM
pipeline. Brackets show paired 95\% bootstrap confidence intervals over
prompts.}
\label{tab:reduced-step-control}
\scriptsize
\setlength{\tabcolsep}{5pt}
\renewcommand{\arraystretch}{1.16}
\resizebox{\textwidth}{!}{%
\begin{tabular}{lcccccc}
\toprule
Configuration & Prompts & Calls & Cached PSNR & Anchor-only PSNR &
Cached $-$ anchor-only & Default three-step PSNR \\
\midrule
FLUX.1-schnell, Euler & 50 & 3 & 32.575 & 32.577 & $-0.001$ [-0.004, 0.001] & \textit{N/A} \\
PixArt-LCM & 50 & 3 & 23.930 & 23.914 & $+0.016$ [0.013, 0.019] & 17.502 \\
\bottomrule
\end{tabular}
}
\end{table}

\begin{table}[H]
\centering
\caption{PixArt-LCM baseline sweep on 50 paired prompts. Aggregate PSNR uses
mean RGB MSE and remains finite when some outputs are exact. FCE is the mean
number of full DiT evaluations.}
\label{tab:baseline-sweep}
\scriptsize
\setlength{\tabcolsep}{7pt}
\renewcommand{\arraystretch}{1.14}
\begin{tabular}{lccccc}
\toprule
Method & Threshold & FCE & Aggregate PSNR & SSIM & Latency (s) \\
\midrule
AutoTarget & \textit{N/A} & 3.000 & \textbf{22.952} & \textbf{0.777} & 0.123 \\
TeaCache & 0.10 & 4.000 & exact & 1.000 & 0.157 \\
TeaCache & 0.30 & 4.000 & exact & 1.000 & 0.158 \\
TeaCache & 0.55 & 3.000 & 6.988 & 0.053 & \textbf{0.121} \\
TeaCache & 0.80 & 3.000 & 5.766 & 0.037 & 0.122 \\
DiCache & 0.10 & 4.000 & exact & 1.000 & 0.157 \\
DiCache & 0.25 & 4.000 & exact & 1.000 & 0.158 \\
DiCache & 0.391 & 3.286 & 7.464 & 0.289 & 0.132 \\
DiCache & 0.60 & 3.036 & 5.766 & 0.037 & 0.123 \\
\bottomrule
\end{tabular}
\end{table}

\begin{table}[H]
\centering
\caption{Solver comparison on the same PixArt-XL model at $1024^2$. Both
solvers use 20 steps, the same 20 prompts and paired seeds, and ten full DiT
evaluations at the same even-numbered steps. The table reports candidates even
when they produce non-finite values.}
\label{tab:solver-control}
\scriptsize
\setlength{\tabcolsep}{4.5pt}
\renewcommand{\arraystretch}{1.16}
\begin{tabular}{lcccccc}
\toprule
Solver & Selected & Ranking & $\epsilon$ PSNR & clean PSNR & velocity PSNR & Finite clean \\
\midrule
DDIM & $\epsilon$ & $\epsilon > v > \mathrm{clean}$ & 14.154 & non-finite & 13.998 & 0/20 \\
DPM-Solver++ (order 2) & $\epsilon$ & $\epsilon > \mathrm{clean} > v$ & 18.145 & 17.936 & 17.760 & 20/20 \\
\bottomrule
\end{tabular}
\end{table}